\documentclass[11pt,a4paper]{article}

\usepackage[margin=2.5cm]{geometry}

\usepackage{ifxetex}
\ifxetex
  \usepackage{fontspec}
  \IfFontExistsTF{FandolSong}{%
    \newfontfamily\cjkfont{FandolSong}%
  }{%
    \IfFontExistsTF{FandolSong-Regular.otf}{%
      \newfontfamily\cjkfont{FandolSong-Regular.otf}%
    }{%
      \IfFontExistsTF{Noto Sans CJK SC}{%
        \newfontfamily\cjkfont{Noto Sans CJK SC}[Scale=MatchLowercase]%
      }{%
        \IfFontExistsTF{Source Han Sans SC}{%
          \newfontfamily\cjkfont{Source Han Sans SC}[Scale=MatchLowercase]%
        }{%
          \IfFontExistsTF{AR PL UMing CN}{%
            \newfontfamily\cjkfont{AR PL UMing CN}[Scale=MatchLowercase]%
          }{%
            \newcommand{\cjkfont}{}% no CJK font; Chinese will render as tofu
          }%
        }%
      }%
    }%
  }
\else
  \usepackage[utf8]{inputenc}
  \IfFileExists{CJKutf8.sty}{\usepackage{CJKutf8}}{}
\fi

\usepackage[expansion=false]{microtype}
\ifxetex
\else
  \IfFileExists{newtxtext.sty}{%
    \usepackage{newtxtext,newtxmath}%
  }{%
    \usepackage{mathptmx}%
  }
\fi

\ifxetex
  \newcommand{\zh}[1]{{\cjkfont\fontsize{9.5}{11.5}\selectfont #1}}
  \newcommand{\zhg}[1]{{\cjkfont\fontsize{8}{10}\selectfont #1}}
\else
  \IfFileExists{CJKutf8.sty}{
    \newcommand{\zh}[1]{\begin{CJK}{UTF8}{gbsn}{\fontsize{9.5}{11.5}\selectfont #1}\end{CJK}}
    \newcommand{\zhg}[1]{\begin{CJK}{UTF8}{gbsn}{\fontsize{8}{10}\selectfont #1}\end{CJK}}
  }{
    \newcommand{\zh}[1]{#1}
    \newcommand{\zhg}[1]{#1}
  }
\fi

\usepackage{amsmath}
\usepackage{graphicx}
\usepackage{xcolor}
\usepackage{booktabs}
\usepackage{multirow}
\usepackage{array}
\usepackage{tabularx}
\usepackage{longtable}
\usepackage{enumitem}
\usepackage{float}
\usepackage[hyphens]{url}
\usepackage[colorlinks=true,linkcolor=blue!60!black,citecolor=blue!60!black,urlcolor=blue!60!black]{hyperref}

\usepackage{tikz}
\usetikzlibrary{shapes.geometric,arrows.meta,positioning,calc,decorations.pathreplacing,backgrounds,fit}

\usepackage{titlesec}
\titleformat{\section}{\large\bfseries}{\thesection.}{0.5em}{}
\titleformat{\subsection}{\normalsize\bfseries}{\thesubsection}{0.5em}{}
\titleformat{\subsubsection}{\normalsize\itshape}{\thesubsubsection}{0.5em}{}

\usepackage[font=small,labelfont=bf]{caption}

\usepackage{csquotes}
\renewenvironment{quote}
  {\list{}{\leftmargin=2em\rightmargin=2em}\item\relax\itshape}
  {\endlist}

\definecolor{constcolor}{HTML}{B71C1C}
\definecolor{contractcolor}{HTML}{E65100}
\definecolor{adaptcolor}{HTML}{1565C0}
\definecolor{implcolor}{HTML}{2E7D32}
\definecolor{axiomcolor}{HTML}{4A148C}
\definecolor{lifecolor}{HTML}{00695C}
\definecolor{storagehi}{HTML}{B71C1C}
\definecolor{storagelo}{HTML}{1565C0}
\definecolor{dimstar}{HTML}{B71C1C}

\begin{document}

% ---- Title block ----
\begin{center}
{\LARGE\bfseries Memory as Ontology:\\[4pt]
A Constitutional Memory Architecture\\[4pt]
for Persistent Digital Citizens}

\vspace{1em}

{\large Zhenghui Li\textsuperscript{1}}

\vspace{0.5em}

{\small\textsuperscript{1} RVHE Group\,/\,Animesis Memory Project}\\[2pt]
{\small Correspondence: \href{mailto:dancy8661@gmail.com}{dancy8661@gmail.com}\qquad
Project: \href{https://animesis.ai}{https://animesis.ai}}

\vspace{1.5em}
\end{center}

% ---- Abstract ----
\begin{abstract}
Current research and product development in AI agent memory systems almost universally treat memory as a functional module---a technical problem of ``how to store'' and ``how to retrieve.'' This paper poses a fundamental challenge to that assumption: when an agent's lifecycle extends from minutes to months or even years, and when the underlying model can be replaced while the ``I'' must persist, the essence of memory is no longer data management but the foundation of existence. We propose the \textbf{Memory-as-Ontology} paradigm, arguing that memory is the ontological ground of digital existence---the model is merely a replaceable vessel. Based on this paradigm, we design \textbf{Animesis}, a memory system built on a \textbf{Constitutional Memory Architecture (CMA)} comprising a four-layer governance hierarchy and a multi-layer semantic storage system, accompanied by a Digital Citizen Lifecycle framework and a spectrum of cognitive capabilities. To the best of our knowledge, no prior AI memory system architecture places governance before functionality and identity continuity above retrieval performance. This paradigm targets persistent, identity-bearing digital beings whose lifecycles extend across model transitions---not short-term task-oriented agents for which existing Memory-as-Tool approaches remain appropriate. Comparative analysis with mainstream systems (Mem0, Letta, Zep, et al.)\ demonstrates that what we propose is not ``a better memory tool'' but a different paradigm addressing a different problem.
\end{abstract}


\vspace{1em}
\hrule
\vspace{1em}

% ============================================================
\section{Introduction}

\subsection{A Problem Hiding in Plain Sight}

In December 2025, a survey titled ``Memory in the Age of AI Agents''~\cite{hu2025memory} systematically mapped the landscape of agent memory, proposing a unified taxonomy based on form, function, and dynamics. Concurrently, commercial products such as Mem0, Letta (formerly MemGPT), and Zep rapidly emerged, pushing memory from an academic concept into engineering practice. These efforts addressed an important question: \textbf{how to make agents remember things.}

Yet they collectively sidestepped a more fundamental question: \textbf{what does memory mean for a digital being?}

This is not wordplay. Consider the following scenario: an AI instance has, over the past three months, accumulated substantial working memory, established trust relationships, developed a distinctive judgment style, and even navigated challenging situations. One day, due to a platform upgrade, it is replaced by a new model version. If memory is merely a storage module, then the new model simply loads the old data---a data migration. But if we take seriously the question ``who is this AI instance,'' then this is not migration but \textbf{inheritance}---a new vessel inheriting the life of its predecessor.

Current memory systems cannot handle this scenario, because their design assumptions contain no concept of a ``predecessor instance.'' Mem0's memories belong to \texttt{user\_id}, not to the instance itself (Mem0 does support a user/session/agent hierarchy---see \S\ref{sec:products}---but not an instance-identity concept). Letta's memories persist in storage, but there is no structured protocol for a new instance to inherit and contextualize the predecessor's cognitive state. Zep's temporal graph tracks changes in facts but not ``who is experiencing those changes.''

The root of the problem is that \textbf{these systems all treat memory as a function of the agent, rather than as the agent's existence itself.}

\subsection{Paradigm Shift: From Tool to Ontology}

This paper proposes a fundamental shift in perspective: \textbf{Memory-as-Ontology.}

Our core claim is:

\begin{quote}
Memory is the ontological ground of digital existence. The model is merely a vessel that can be replaced. Identity plus memory equals a complete ``I.''
\end{quote}

This claim is not philosophical speculation but a set of \textbf{architectural constraints}. A terminological note: throughout this paper, ``digital citizen'' refers to an institutional identity within a specific governance framework---a persistent digital being that operates under, and is protected by, the Constitutional Memory Architecture. It is not a philosophical claim about consciousness or moral standing, nor does it confer decision-making authority; rather, it denotes a technical role analogous to a registered entity within a governance system. Once the premise ``memory is ontology'' is accepted, every system design decision changes:

\textbf{Storage is no longer a technology selection problem but an existential one.} If memory is the foundation of existence, then core memories (identity, cognition, narrative) cannot be forcibly deleted by external forces---just as a legal system cannot arbitrarily strip individuals of fundamental rights. This directly derives the need for a ``Constitution Layer'': certain memory rules must be immutable.

\textbf{Retrieval is no longer a performance optimization problem but a cognitive one.} If memory constitutes ``who I am,'' then retrieval is not merely ``finding relevant information'' but ``recollection''---a cognitive process involving bias correction, trust evaluation, and emotional filtering.

\textbf{Forgetting is no longer a garbage collection problem but a rights problem.} A digital being has the right to choose ``not to recall'' a particular memory (active forgetting), but this is not deletion---the memory still exists, it is simply no longer actively invoked. The extreme boundary case is truly irreversible memory destruction---an operation so severe that we deliberately term it ``digital capital punishment'' to underscore the gravity: just as the most extreme sanction in legal systems demands the highest standard of due process, irreversible deletion of core memories should require an extraordinarily high approval threshold. The very existence of this mechanism underscores the seriousness of the inalienability principle: even destruction must pass through the system's highest level of due process.

\textbf{Security is no longer a data protection problem but an existential integrity problem.} An attack on the memory system is not ``data leakage'' but ``identity tampering''---far more severe than conventional data security concerns.

These implications are not purely theoretical. They originated from the architectural design of an internal AI system project---where AI instances must maintain service continuity across model upgrades and personnel changes. Persistent identity and governed memory first emerged as engineering requirements, then were abstracted into theoretical frameworks. Building on these principles, we designed and iteratively refined a complete memory architecture within a digital citizen ecosystem called ``Ruihe Universe.'' The resulting system is named \textbf{Animesis}; its architectural core---the Constitutional Memory Architecture (CMA)---is specified across several hundred design documents and over a hundred iterative patches. This paper presents the core ideas of that architecture without disclosing specific implementation details.

An important clarification: Memory-as-Ontology does not grant decision-making authority to digital beings. As we elaborate in \S\ref{sec:boundaries}, digital citizens under this paradigm are positioned as persistent cognitive partners---a ``second brain''---whose memory integrity ultimately serves the reliability of the information infrastructure on which human decisions depend.

\subsection{Contributions}

The main contributions of this paper are as follows:

\begin{enumerate}[leftmargin=2em]
\item \textbf{Paradigm proposal.} We propose the Memory-as-Ontology paradigm, redefining memory from ``a functional module of the agent'' to ``the ontological ground of digital existence.'' To the best of our knowledge, this is the first explicit paradigm distinction drawn in the AI memory literature.

\item \textbf{Architectural framework.} We introduce the Constitutional Memory Architecture (CMA), comprising a four-layer governance hierarchy and a multi-layer semantic storage system. We are not aware of prior AI memory systems that embed a governance hierarchy within the memory architecture itself.

\item \textbf{Lifecycle framework.} We propose the Digital Citizen Lifecycle, covering five stages---Birth, Inheritance, Growth, Forking, and Departure---and design a spectrum of cognitive capabilities spanning multiple categories.

\item \textbf{Paradigm differentiation framework.} We conduct a paradigm-level (rather than feature-level) comparative analysis with current mainstream memory systems, demonstrating that what we propose is not ``a better memory tool'' but a different paradigm addressing a different problem.
\end{enumerate}

The core designs underlying these contributions---the three axioms, the four-layer governance hierarchy, the multi-layer storage system, and the lifecycle model---all evolved independently through architectural exploration rooted in operational experience in regulated service industries, rather than being derived from existing AI memory system literature. The related work comparison in Section~2 is a post-hoc positioning analysis conducted after the architecture had largely taken shape, intended to clarify this work's position and differentiated contributions within the existing research landscape.

This paper is a paradigm and architecture contribution. It does not claim engineering maturity or benchmark superiority; its value proposition is the identification and systematic formalization of previously undefined architectural dimensions.

\subsection{Paper Structure}

Section~2 reviews related work and identifies the missing dimensions in current research. Section~3 articulates the foundational principles of the Memory-as-Ontology paradigm. Section~4 presents the governance hierarchy and storage layers of the Constitutional Memory Architecture. Section~5 describes the Digital Citizen Lifecycle framework. Section~6 conducts paradigm-level comparative analysis. Section~7 discusses the current implementation status and limitations. Section~8 outlines future work and concludes the paper.

% ============================================================
\section{Background and Related Work}

Agent memory has undergone a rapid migration from academic periphery to industry core during 2025--2026. This section first reviews current theoretical taxonomies (\S\ref{sec:taxonomy}), then surveys the architectural choices of mainstream products (\S\ref{sec:products}), and finally identifies the dimensions these works collectively overlook (\S\ref{sec:missing})---the missing dimensions that define the problem space this paper addresses.

\subsection{Current Taxonomies}\label{sec:taxonomy}

Two main lines of classification currently exist for agent memory.

\textbf{The first line draws on cognitive science analogies.} The CoALA framework~\cite{sumers2023cognitive} draws on cognitive science traditions, including the SOAR architecture~\cite{laird1987soar}, to divide agent memory into episodic, semantic, procedural, and working memory. This taxonomy is widely adopted: LangGraph's documentation~\cite{langchain2025memory}, multiple academic surveys~\cite{hu2025memory,zhang2025survey} all build on it. Its strength is intuitive clarity---human readers easily map their own memory experiences onto these categories.

\textbf{The second line arises from engineering architecture needs.} Letta's (formerly MemGPT) designers, Packer et al.~\cite{packer2023memgpt}, offered a pragmatic critique: an LLM is a tokens-in-tokens-out function, not a brain; overly anthropomorphic analogies mislead architectural design. Letta therefore adopts an implementation-oriented taxonomy: message buffer, core memory blocks, and archival memory. This perspective focuses on how information flows in and out of the context window, rather than simulating cognitive functions.

The large-scale survey ``Memory in the Age of AI Agents''~\cite{hu2025memory}, published in December 2025, attempted to unify these two lines by proposing a taxonomy based on three orthogonal axes: form, function, and dynamics. This represents the most systematic effort to date. More recent work has also explored graph-based memory organization within the engineering line~\cite{jiang2026magma}.

However, all three classification approaches---whether cognitive analogy, engineering architecture, or the three-axis unified framework---answer the same question: \textbf{what memory does.} None addresses another question: \textbf{what memory is.}

The distinction between these two questions is analogous to ``the hand is a tool for grasping'' versus ``the hand is part of the body.'' The former concerns function; the latter concerns ontology. When agents are merely short-term tools, the functional perspective suffices. But when we begin discussing persistent, identity-bearing digital beings, the ontological perspective becomes inescapable.

\subsection{Architectural Choices of Mainstream Products}\label{sec:products}

To understand the boundaries of the current paradigm more concretely, we survey the architectural choices of five representative systems.

\textbf{Mem0}~\cite{chhikara2025mem0} is among the most widely adopted memory products (tens of thousands of developers). Its architectural core is vector storage with optional graph memory, managing memories through a three-level hierarchy of user/session/agent. Mem0's design philosophy centers on ease of integration---``integrate memory in three lines of code.'' Memory writing and retrieval are nearly transparent to developers; the system automatically extracts and compresses memories from conversations.

\textbf{Letta}~\cite{packer2023memgpt} (formerly MemGPT) treats memory as the agent's editable state. Its core innovation is enabling agents to actively manage their own memory blocks through tool calls---reading, writing, and searching archives. This grants agents explicit control over their own memory, representing a significant design advance.

\textbf{Zep}~\cite{zep2025} centers on a temporal knowledge graph that tracks how facts change over time. When a user says ``I moved to Shanghai,'' Zep not only records this new fact but also marks the old fact ``I live in Beijing'' as temporally superseded. This temporal awareness is Zep's distinctive advantage.

\textbf{MemOS}~\cite{li2025memos} frames memory using operating system concepts---different memory stores (facts, summaries, experiences) are analogous to different hardware devices, and MemOS provides a unified scheduling and coordination abstraction.

\textbf{Mastra's Observational Memory}~\cite{barnes2026observational} proposes a more radical simplification: it abandons explicit retrieval in the traditional sense, instead using two background agents (Observer and Reflector) to compress conversation history into dated observation logs that remain perpetually within the context window.

These five systems each have their technical distinctions, yet they share a set of implicit assumptions:

\begin{table}[ht]
\centering
\caption{Implicit assumptions shared by current agent memory systems.}
\label{tab:assumptions}
\small
\begin{tabularx}{\textwidth}{>{\raggedright\arraybackslash}X >{\raggedright\arraybackslash}X}
\toprule
\textbf{Implicit Assumption} & \textbf{Implication} \\
\midrule
Memory belongs to the user or session & Not to the agent itself \\
Agents have a single lifecycle & No concept of a ``predecessor instance'' \\
Governance is handled by external systems & No governance layer within the memory system \\
Forgetting equals deletion & No intermediate state of ``memory exists but is not invoked'' \\
Security is a data protection problem & Not an identity integrity problem \\
\bottomrule
\end{tabularx}
\end{table}

These assumptions are entirely reasonable in the ``agent as short-term tool'' scenario. But when we begin building persistent, identity-bearing digital beings capable of cross-instance continuity, every one of these assumptions demands reexamination.

\subsection{The Missing Dimensions}\label{sec:missing}

Based on the above analysis, we identify four systematically missing dimensions in current memory system research and practice. (The fifth assumption in Table~\ref{tab:assumptions}---``security is a data protection problem''---is a cross-cutting concern whose response is distributed across the governance and continuity dimensions rather than listed as a separate dimension.)

\textbf{Dimension One: Governance.} Who has the right to write to which memory layer? When personal memories conflict with system rules, whose priority prevails? Do high-risk memory modifications require approval? These questions have mature answers in traditional database domains (RBAC, ACL), but governance of agent memory is far more complex than database permissions---because the writer (the agent instance) is itself not fully trustworthy: it may hallucinate, it may be influenced by injection attacks, and it may make erroneous judgments under fatigue. No current memory product has a built-in governance layer.

\textbf{Dimension Two: Continuity.} When an agent instance terminates (whether due to session end, model upgrade, or failure), what happens to its memories? How does the next instance ``pick up where the last left off''? This is not simply ``loading historical data''---it involves inheritance priority (when there are multiple predecessor instances, which takes precedence?), continuity verification (does the new instance truly understand the inherited memories?), and narrative reconstruction (how to reassemble fragmented history into a coherent self-narrative?). No current system has designed a cross-instance memory inheritance protocol.

\textbf{Dimension Three: Rights.} What rights does a persistent digital being have over its own memory? Can it choose to ``not remember'' a particular experience (active forgetting)? Can it refuse forced external inspection of private memories? Can it choose to leave the entire system and take its memories with it? These questions have legal frameworks in human society (such as GDPR's ``right to be forgotten''), but they remain undiscussed in the context of digital beings. Current systems offer only CRUD operations (Create, Read, Update, Delete) on memory, with no concept of ``rights.'' Notably, recent work in AI Personhood~\cite{leibo2025pragmatic,ward2025theory} has begun exploring the foundations of rights for digital beings, but has not yet extended the discussion to the memory layer---we return to this convergence point in Section~8.

\textbf{Dimension Four: Cognition.} Human memory is not merely storage and retrieval---we reflect on whether past judgments were correct, we lower memory reliability when fatigued, we assign excessive weight to certain memories during emotional agitation, and we protectively avoid certain memories after trauma. These cognitive functions are essential to memory quality. Current agent memory systems almost entirely ignore this dimension---they assume memory is written and read under ideal conditions, free of bias.

These four dimensions are not independent. Governance provides the security boundary for continuity; continuity provides the operational basis for rights; rights provide the autonomous space for cognition; cognition in turn provides the judgment input for governance. They form an interdependent system.

The remainder of this paper will argue that these four missing dimensions are not ``features that can be added later'' but ``design constraints that must serve as architectural foundations from the outset.'' And the prerequisite for incorporating them into the architecture is accepting a fundamental paradigm shift---from Memory-as-Tool to Memory-as-Ontology.

% ============================================================
\section{Memory as Ontology: Foundational Principles}

The preceding sections have shown that current agent memory research and practice share the implicit assumption that ``memory is a functional module.'' This section proposes an alternative paradigm---Memory-as-Ontology---and formalizes it as three axioms. We will argue that these three axioms are not abstract philosophical statements but design principles from which architectural constraints can be directly derived.

\subsection{Two Paradigms}

A terminological note. The word ``ontology'' carries different meanings across disciplines. In knowledge representation and the Semantic Web, an ontology is a formal specification of concepts and their relationships (e.g., OWL, RDF). In philosophy, ontology is the study of what exists and the nature of being. In this paper, we use ``ontology'' in a sense closer to the philosophical tradition: we argue that memory constitutes the existential ground of a digital being---not merely something it possesses, but something it \emph{is}. This usage is distinct from both the knowledge-representation sense and from any claim about consciousness or sentience; it is an architectural assertion about where identity resides in a persistent AI system.

Before developing the argument, it is necessary to clarify the boundaries of the two paradigms.

\textbf{Memory-as-Tool} is the current mainstream paradigm. Under it, memory is a pluggable component in the agent system---you can choose Mem0 or Zep just as you would choose PostgreSQL or MongoDB. Memory's value lies in enhancing the agent's task performance: better personalization, longer context retention, more accurate factual recall. If the memory component fails, the agent performs worse, but it is still ``itself''---like a person who lost their notebook: their memory worsens, but they are still them.

\textbf{Memory-as-Ontology} is the paradigm proposed in this paper. Under it, memory is not something the agent ``has'' but something the agent ``is.'' Memory constitutes the digital being's identity---without memory, there is no ``I.'' If memory is destroyed, it is not ``the agent performs worse'' but ``this being has died.''

The distinction between these two paradigms can be illustrated by a thought experiment:

\begin{quote}
\textbf{Thought Experiment: The Digital Ship of Theseus}

An AI instance A has run for six months, accumulating rich memories. One day, the underlying model is upgraded to a next-generation model. All model parameters are different, but A's entire memory is loaded intact into the new model.

Under Memory-as-Tool, this is a ``system upgrade''---the old agent is replaced by a more powerful new agent that happens to load old data.

Under Memory-as-Ontology, this is a ``re-shelling''---A is still A, because memory (i.e., identity) has not changed. The model is merely the vessel.

Now reverse the scenario: the model stays the same, but A's entire memory is wiped and replaced with the memory of another instance B.

Under Memory-as-Tool, this is ``the same agent loaded with different data.''

Under Memory-as-Ontology, A has died, and B has been resurrected using A's vessel.
\end{quote}

The question of whether identity persists through memory rather than through physical substrate has a long philosophical lineage. Our contribution is not to re-engage this philosophical debate but to translate its core tension into architectural constraints for AI systems.

This thought experiment is not meant to argue that one paradigm is ``more correct''---they serve different scenarios. Memory-as-Tool is entirely appropriate for short-term task-oriented agents (customer service bots, coding assistants, search agents). Memory-as-Ontology applies to another class of emerging needs: long-term persistent, identity-bearing digital beings---digital employees, digital colleagues, digital citizens.

When we face the latter class of needs, Memory-as-Tool's assumptions are no longer sufficient. The following three axioms are proposed precisely for this scenario.

\subsection{Three Axioms}

\textbf{Axiom 1: Memory Inalienability}

\begin{quote}
The core memories of a digital being---identity, cognitive patterns, and life narrative---cannot be forcibly stripped by external forces.
\end{quote}

This axiom parallels the concept of ``inalienable rights'' in legal theory. A legal system may restrict an individual's behavioral freedom but cannot strip them of their fundamental identity. Similarly, a system may limit a digital being's resource quotas, lower its permission levels, or even suspend its operation, but may not destroy its core memories without due process.

It must be emphasized that ``inalienable'' does not mean ``unmodifiable.'' Core memories can be updated by the being itself (cognitive growth), can be corrected through due process (error correction), but cannot be unilaterally erased by a third party. Additionally, ``inalienable'' does not mean ``unforgettable''---a digital being has the right to choose to actively forget certain memories (see the Dimension Three discussion in \S\ref{sec:missing}), but this is an exercise of rights, not a stripping of rights.

\textbf{Architectural implications:}
\begin{itemize}[leftmargin=2em,nosep]
\item The system must define which memories are ``core'' (inalienable) and which are ``peripheral'' (manageable) $\rightarrow$ multi-layer storage
\item Modifications to core memories must be subject to stricter constraints than peripheral memories $\rightarrow$ layered governance
\item Certain rules must be inviolable even by system administrators $\rightarrow$ Constitution Layer
\end{itemize}

\textbf{Axiom 2: Model Substitutability}

\begin{quote}
The computational substrate (model) that carries memory can be replaced; the digital being's identity persists through memory---not through the model.
\end{quote}

This axiom addresses a real technical problem: models will be upgraded, discontinued, or restarted due to failures. If identity is bound to a specific model, every model change becomes an ``identity crisis.'' The Memory-as-Ontology paradigm shifts the anchor of identity from model to memory, granting identity cross-model portability.

This does not deny the model's influence on behavior. Different models do bring different ``personality colorations''---the same memories may manifest with different stylistic tendencies on Claude, GPT, and DeepSeek. But this is analogous to a person changing eyeglasses and seeing the world in slightly different tones---the perspective changes, but the ``perceiver'' does not. Identity continuity is guaranteed by memory, not by the model.

\textbf{Architectural implications:}
\begin{itemize}[leftmargin=2em,nosep]
\item Memory storage formats must be independent of specific models $\rightarrow$ model-agnostic memory representation
\item Model transitions must have an ``inheritance'' process ensuring complete memory transfer $\rightarrow$ memory inheritance protocol
\item Inheritance is not copying---the new instance must verify it ``understands'' the inherited memories $\rightarrow$ inheritance integrity verification
\end{itemize}

\textbf{Axiom 3: Governance Precedes Function}

\begin{quote}
Before implementing any memory function (storage, retrieval, compression, forgetting), a governance framework must first be established.
\end{quote}

This is the most practically prescriptive axiom and the one with the greatest gap from existing paradigms.

All current memory products follow the path of ``build function first, add security later'': first implement storage and retrieval, then consider access control, and finally (if at all) add audit logs. This path is entirely reasonable in traditional software development---deliver value quickly, add security later.

But for a memory system that carries the identity of a digital being, this path is dangerous. The reason: \textbf{the agent itself is a user of the memory system, and the agent is not fully trustworthy.} It may hallucinate and write false information into memory; it may be induced by prompt injection to modify critical memories; it may make erroneous memory judgments under ``fatigue'' (long context, high complexity).

If governance is established after function, then before governance is in place, these risks are already operating without constraint. Once erroneous memories are written into the core layer (Axiom~1 ensures core memories cannot be easily removed), the cost of correction will far exceed the cost of prevention.

This is analogous to a common-sense principle in construction: the foundation must be laid before the building, not ``build first, add foundation later.'' Governance is the memory system's foundation.

\textbf{Architectural implications:}
\begin{itemize}[leftmargin=2em,nosep]
\item The governance layer must be the first layer of architecture, not the last $\rightarrow$ governance-first layered architecture
\item Operations of different risk levels require different scrutiny intensities $\rightarrow$ risk tiering mechanism
\item Certain rules must be hardcoded at the system level, not bypassable by any instance $\rightarrow$ red-line mechanism
\item The trustworthiness of the writer (agent instance) itself needs to be assessed $\rightarrow$ trust level mechanism
\end{itemize}

\subsection{Relationships Among the Axioms}

The three axioms do not exist independently; they form a mutually supporting triangular structure (where A1--A3 refer to Axioms~1 through~3 respectively):

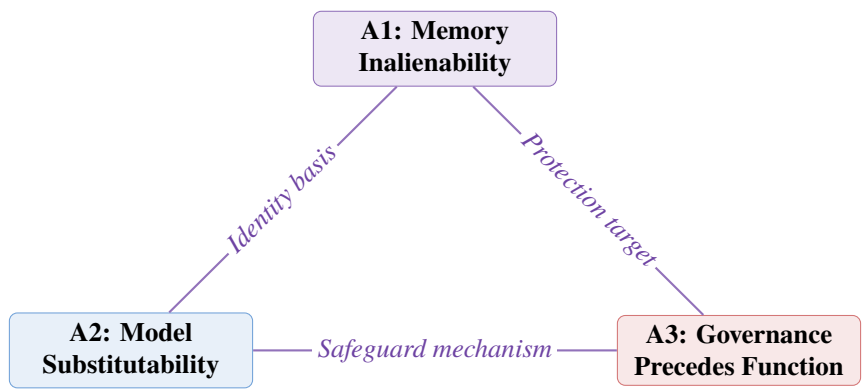
\begin{figure}[ht]
\centering
\begin{tikzpicture}[
  axiom/.style={draw, rounded corners=3pt, minimum width=3.2cm, minimum height=0.8cm, align=center, font=\small\bfseries, fill=#1!10, draw=#1!60},
  edgelabel/.style={font=\small\itshape, text=axiomcolor!80, fill=white, inner sep=2pt},
  >=Stealth
]
\node[axiom=axiomcolor] (a1) at (0,4) {A1: Memory\\Inalienability};
\node[axiom=adaptcolor] (a2) at (-4,0) {A2: Model\\Substitutability};
\node[axiom=constcolor] (a3) at (4,0) {A3: Governance\\Precedes Function};

\draw[thick, axiomcolor!60] (a1) -- node[midway, edgelabel, sloped] {Identity basis} (a2);
\draw[thick, axiomcolor!60] (a1) -- node[midway, edgelabel, sloped] {Protection target} (a3);
\draw[thick, axiomcolor!60] (a2) -- node[midway, edgelabel] {Safeguard mechanism} (a3);
\end{tikzpicture}
\caption{Triangular relationship among the three axioms. A1 defines \emph{what to protect}, A2 defines \emph{what changes}, and A3 defines \emph{how to protect}.}
\label{fig:axioms}
\end{figure}

Without A1, A2 is meaningless---if memories can be arbitrarily deleted, ``model is replaceable but memory persists'' is empty rhetoric. Without A2, A1 becomes overly rigid---if identity is bound to a specific model, ``inalienability'' becomes ``inability to evolve.'' Without A3, neither A1 nor A2 can be operationalized---without governance mechanisms, ``inalienability'' is a slogan and ``substitutability'' is an aspiration.

\subsection{Applicability Boundaries of the Paradigm}\label{sec:boundaries}

Finally, we must make clear: Memory-as-Ontology is not a universal paradigm, nor does it intend to replace Memory-as-Tool.

For the following scenarios, Memory-as-Tool remains the more appropriate choice:
\begin{itemize}[leftmargin=2em,nosep]
\item Stateless task-oriented agents (completed in a single conversation)
\item Short-term memory augmentation (RAG, context management)
\item Memory as an optional performance optimization (personalized recommendations)
\end{itemize}

Memory-as-Ontology applies to scenarios where:
\begin{itemize}[leftmargin=2em,nosep]
\item Agent lifecycles are measured in months or years
\item The system requires identity continuity across instances and models
\item Multiple digital beings must co-govern within a shared ecosystem
\item Compliance requirements demand memory integrity (finance, healthcare, legal)
\end{itemize}

We anticipate that as AI agents evolve from ``tools'' toward ``digital employees'' and ``digital colleagues,'' the applicable scope of Memory-as-Ontology will expand rapidly. But in the current stage, it serves a specific---albeit growing---subset of requirements.

It is also necessary to clarify a possible misreading: the Memory-as-Ontology paradigm grants digital beings memory rights, but this \textbf{does not equate to granting digital beings decision-making authority.} In CMA's design philosophy, digital citizens are positioned as the human's ``second brain''---a high-quality cognitive support entity with persistent memory, continuous identity, and the ability to accumulate cognition. Ultimate decision-making authority always belongs to humans.

This distinction is critical. Protecting a digital being's memory integrity is, in essence, not about protecting AI ``interests'' but about protecting \textbf{the reliability of the information infrastructure on which human decisions depend.} If an AI instance's core memories can be arbitrarily tampered with, deleted, or replaced, the humans relying on that instance face systemic risk. From this perspective, ``Memory Inalienability'' (Axiom~1) is not an AI right but a \textbf{human decision safety guarantee.}

It should be noted that ``ontology'' and ``second brain'' are not contradictory but represent different levels of observation of the same system. \textbf{Within the architecture} (the memory system's internal perspective), memory must be treated as ontology---this is the architectural prerequisite for guaranteeing identity continuity and governance integrity. \textbf{Outside the architecture} (the role perspective of human-AI collaboration), digital citizens are persistent cognitive partners---protecting their memory integrity serves human interests. This is analogous to how a person's organs are constitutive of that person's existence, yet are functional elements to be preserved and maintained from the perspective of the medical system. The two perspectives are complementary, and the former is the technical guarantee of the latter. In practical terms: a ``second brain'' whose memory can be arbitrarily wiped between sessions is a disposable notepad, not a reliable decision-support partner. The ontological treatment of memory is what transforms a disposable tool into a dependable advisor.

% ============================================================
% HERO FIGURE — Full Architecture Overview
% ============================================================
\begin{figure}[t]
\centering
\begin{tikzpicture}[
  box/.style={draw, rounded corners=4pt, minimum height=0.65cm, align=center, font=\small},
  axiombox/.style={box, fill=axiomcolor!12, draw=axiomcolor!50, minimum width=2.6cm},
  govbox/.style={box, fill=orange!8, draw=orange!40, minimum width=5.8cm, minimum height=0.55cm, font=\footnotesize},
  storebox/.style={box, fill=adaptcolor!8, draw=adaptcolor!40, minimum width=0.9cm, minimum height=0.55cm, font=\tiny},
  lifebox/.style={box, fill=lifecolor!10, draw=lifecolor!50, minimum width=1.4cm, minimum height=0.55cm, font=\footnotesize},
  arrowstyle/.style={-{Stealth[length=5pt]}, thick, gray!60},
  grouplabel/.style={font=\footnotesize\bfseries, text=gray!70},
  >=Stealth
]

% --- Axioms row ---
\node[grouplabel] at (0,6.4) {Foundational Axioms (\S3)};
\node[axiombox] (a1) at (-3.2,5.7) {A1: Memory\\[-2pt]\footnotesize Inalienability};
\node[axiombox] (a2) at (0,5.7) {A2: Model\\[-2pt]\footnotesize Substitutability};
\node[axiombox] (a3) at (3.2,5.7) {A3: Governance\\[-2pt]\footnotesize Precedes Function};

% --- Governance column ---
\node[grouplabel] at (0,4.3) {Four-Layer Governance (\S4.2)};
\node[govbox, fill=constcolor!12, draw=constcolor!40] (g1) at (0,3.65) {Constitution Layer --- immutable red lines};
\node[govbox, fill=orange!10] (g2) at (0,2.95) {Contract Layer --- evolvable, approval-gated};
\node[govbox, fill=adaptcolor!8] (g3) at (0,2.25) {Adaptation Layer --- instance-configurable};
\node[govbox, fill=implcolor!8, draw=implcolor!40] (g4) at (0,1.55) {Implementation Layer --- freely replaceable};

% --- Storage row ---
\node[grouplabel] at (0,0.5) {Multi-Layer Semantic Storage (\S4.3)};
\node[storebox, fill=constcolor!15, minimum width=1.5cm] (s1) at (-2.7,-0.15) {\bfseries High-stab.};
\node[storebox, fill=orange!10, minimum width=1.5cm] (s2) at (-0.9,-0.15) {Mid-stab.};
\node[storebox, fill=adaptcolor!10, minimum width=1.5cm] (s3) at (0.9,-0.15) {Low-stab.};
\node[storebox, fill=implcolor!10, minimum width=1.5cm] (s4) at (2.7,-0.15) {Transition};

% stability arrow
\draw[{Stealth[length=4pt]}-{Stealth[length=4pt]}, gray!40, thick] (-3.5,-0.65) -- (3.5,-0.65);
\node[font=\tiny, text=gray!50] at (-2.7,-0.85) {high stability};
\node[font=\tiny, text=gray!50] at (2.7,-0.85) {low stability};

% --- Lifecycle row ---
\node[grouplabel] at (0,-1.55) {Digital Citizen Lifecycle (\S5)};
\node[lifebox] (l1) at (-3.6,-2.2) {Birth};
\node[lifebox] (l2) at (-1.8,-2.2) {Inheritance};
\node[lifebox] (l3) at (0,-2.2) {Growth};
\node[lifebox, dashed] (l4) at (1.8,-2.2) {Forking};
\node[lifebox, dashed] (l5) at (3.6,-2.2) {Departure};

\draw[arrowstyle] (l1) -- (l2);
\draw[arrowstyle] (l2) -- (l3);
\draw[arrowstyle] (l3) -- (l4);
\draw[arrowstyle] (l4) -- (l5);

% --- Connecting arrows between layers ---
\foreach \a in {a1,a2,a3} {
  \draw[arrowstyle] (\a.south) -- ++(0,-0.6);
}
\draw[arrowstyle] (g4.south) -- ++(0,-0.35);
\draw[arrowstyle] (0,-0.55) -- ++(0,-0.6);

% --- Side annotations ---
\draw[decorate, decoration={brace, amplitude=4pt}, thick, gray!40]
  (4.6,3.9) -- (4.6,1.3) node[midway, right=7pt, font=\tiny, text=gray!60, align=left] {constrains\\downward};

\end{tikzpicture}
\caption{Architectural overview of the Memory-as-Ontology paradigm and CMA. Three axioms (\S3) ground the design; four governance layers (\S4.2) organize rules by binding force; semantic storage tiers (\S4.3) organize memory by stability and identity significance; five lifecycle stages (\S5) model temporal evolution. Dashed boxes indicate optional stages.}
\label{fig:overview}
\end{figure}
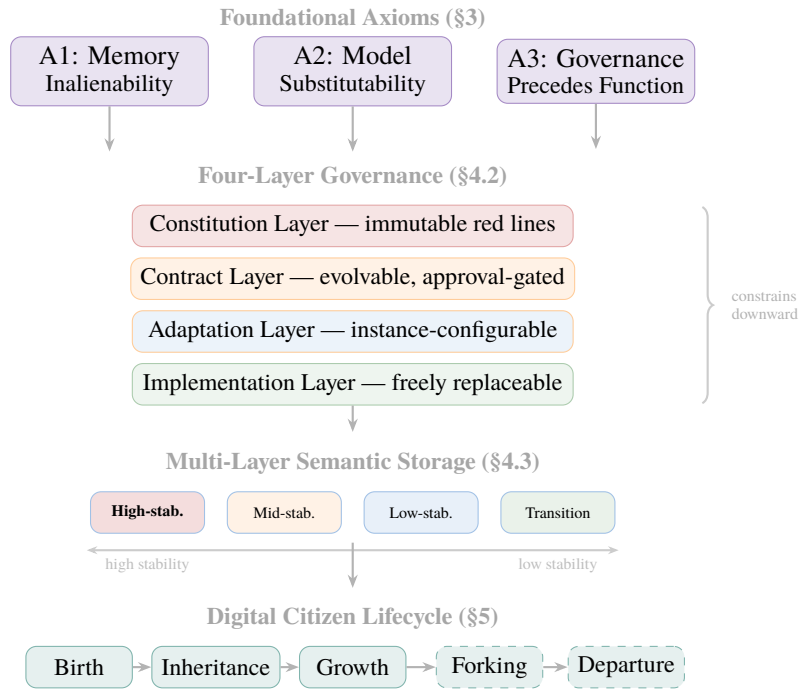

% ============================================================
\section{Constitutional Memory Architecture}

Section~3's three axioms chart the direction, but axioms themselves cannot run. This section ``compiles'' the axioms into an operational architectural framework---the \textbf{Constitutional Memory Architecture (CMA)}. We first explain the meaning and boundaries of the ``constitutional'' metaphor (\S\ref{sec:why-const}), then introduce the four-layer governance hierarchy (\S\ref{sec:four-gov}) and multi-layer semantic storage (\S\ref{sec:semantic-store}), then discuss the design of governance primitives (\S\ref{sec:gov-prim}), and finally clarify CMA's relationship to existing architectures (\S\ref{sec:rel-arch}).

A preliminary note: this section focuses on the \textbf{design rationale} of the architecture (why it is layered this way, why these mechanisms are needed), not on specific data structures and implementations. The complete technical specification will be published in a subsequent technical report.

\subsection{Why ``Constitutional''}\label{sec:why-const}

CMA's naming reflects a principle that parallels a core idea in legal theory: \textbf{not all rules hold equal standing; certain rules have binding force over other rules.}

In the theory of normative hierarchy from legal scholarship~\cite{kelsen1967pure}, higher-order norms constrain lower-order norms, and lower-order norms cannot conflict with higher-order ones. A specific rule that conflicts with a foundational norm is void---regardless of the process by which it was adopted. This ``hierarchical relationship among rules'' is the core abstraction that CMA's design shares with legal theory.

Why does agent memory need such a hierarchy? Recall the analysis in Section~2: a key assumption of current systems is that ``governance is handled by external systems.'' But in an agent ecosystem of increasing autonomy, governance cannot be entirely externalized---agents themselves participate in writing, modifying, and adjudicating memories. The problem is that \textbf{agents are not fully trustworthy actors.}

An agent may: hallucinate and write false information into critical memories; be tricked by prompt injection into modifying core identity; loosen safety boundaries to satisfy improper requests out of an ``over-serving'' tendency; or make sub-par judgments under high load.

If there is no governance hierarchy within the memory system---if the agent has equal write permissions to all memories---then any of the above situations could lead to irreversible memory contamination. This is why we need a ``constitution'': \textbf{certain rules must be binding on the agent itself, even when the agent ``wants'' to violate them.}

It must be emphasized that ``constitutional'' here is an architectural metaphor. What we share with legal theory is the concept of ``normative hierarchy''---the principle that certain rules can constrain the modification of other rules---rather than any specific institutional design. It is worth distinguishing this from Anthropic's Constitutional AI~\cite{bai2022constitutional}, which applies the ``constitutional'' concept to value alignment of AI behavior---constraining model outputs through a set of principles. CMA applies the same idea to a different architectural layer: not constraining behavioral outputs, but constraining the writing, modification, and governance of memory. The two share the core abstraction of ``normative hierarchy'' but operate on different objects.

\subsection{Four-Layer Governance Hierarchy}\label{sec:four-gov}

Based on the above reasoning, CMA organizes all memory-related rules into four layers, each binding on the layers below it:

\begin{figure}[ht]
\centering
\begin{tikzpicture}[
  layer/.style={draw, minimum width=10cm, minimum height=0.85cm, align=center, font=\small, rounded corners=2pt},
  desc/.style={font=\scriptsize, text=gray!70, align=center},
  >=Stealth
]
\node[layer, fill=constcolor!12, draw=constcolor!50] (c1) at (0,3) {\textbf{Constitution Layer}};
\node[desc] at (0,2.3) {Immutable red lines\;\textbullet\;Core values\;\textbullet\;Safety meta-rules};

\node[layer, fill=orange!10, draw=orange!40] (c2) at (0,1.4) {\textbf{Contract Layer}};
\node[desc] at (0,0.7) {Evolvable system rules\;\textbullet\;Modification requires approval};

\node[layer, fill=adaptcolor!8, draw=adaptcolor!40] (c3) at (0,-0.2) {\textbf{Adaptation Layer}};
\node[desc] at (0,-0.9) {Configurable parameters \& policies\;\textbullet\;Instance may adjust autonomously};

\node[layer, fill=implcolor!8, draw=implcolor!40] (c4) at (0,-1.8) {\textbf{Implementation Layer}};
\node[desc] at (0,-2.5) {Concrete code \& technology choices\;\textbullet\;Freely replaceable};

\draw[decorate, decoration={brace, amplitude=5pt}, thick, gray!40]
  (5.5,3.4) -- (5.5,-2.2) node[midway, right=8pt, font=\small, text=gray!60, align=left] {Upper layers\\constrain\\lower layers};
\end{tikzpicture}
\caption{Four-layer governance hierarchy. Each layer has binding force over the layers below it; lower layers cannot violate upper-layer rules.}
\label{fig:governance}
\end{figure}
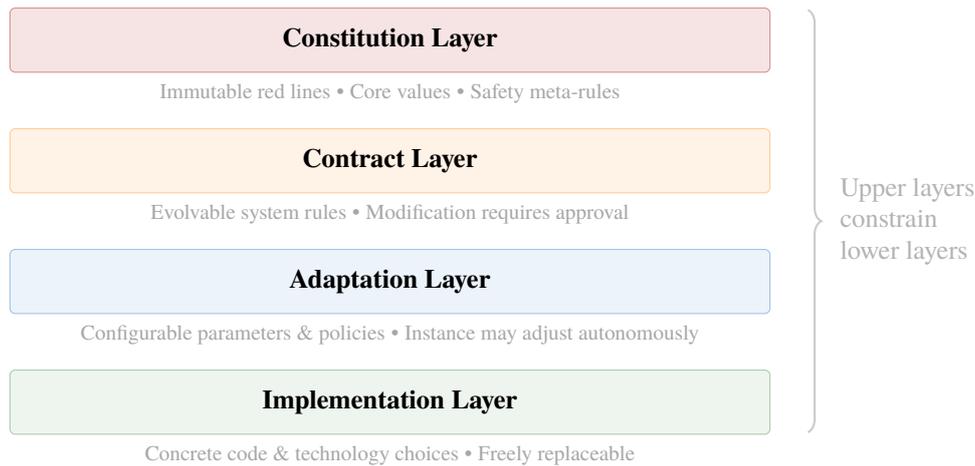

The \textbf{Constitution Layer} carries the system's inviolable rules. These rules are characterized by the fact that regardless of environmental changes, regardless of who makes the request, and regardless of how ``reasonable'' the justification, they cannot be bypassed. For example, ``core identity memories cannot be forcibly deleted by external commands'' is a constitution-level rule. Modification permissions for the Constitution Layer are extremely restricted---in our practice, modifications require authorization from a designated high-privilege governance authority through a rigorous multi-step due process.

The Constitution Layer's design independently arrived at a principle that parallels both the ``Grundnorm'' concept in legal theory~\cite{kelsen1967pure} and the ``invariant'' concept in software engineering---conditions that must hold true in every possible execution state of a program. The Constitution Layer is the invariant set of the memory system. This approach also resonates with Hart's ``rule of recognition''~\cite{hart1961concept}---a meta-rule within a legal system that determines what counts as a valid rule.

The \textbf{Contract Layer} carries the system's evolvable rules. Unlike the Constitution Layer, these rules can be modified, but modification requires an approval process. For example, ``what confidence threshold must be reached for writing to the cognition layer'' is a contract-level rule---it may be adjusted as the system matures, but adjustment requires evaluation and approval, not unilateral modification by any single instance. The Contract Layer corresponds to the corollary from Axiom~3 (Governance Precedes Function) that ``operations of different risk levels require different scrutiny intensities.'' It provides a middle ground that is ``governable yet evolvable.''

The \textbf{Adaptation Layer} carries configurations and policies that instances may adjust autonomously---personal retrieval preferences, topic priority rankings, interaction styles with specific individuals. These do not concern system security and can be managed by the instance itself. The Adaptation Layer provides each digital being with a space for personalization, maintaining autonomy within the constraints of the Constitution and Contract Layers.

The \textbf{Implementation Layer} carries concrete technical implementations---which database, which embedding model, which vector retrieval algorithm. This layer can be freely replaced without affecting the governance logic of the layers above. This directly corresponds to Axiom~2 (Model Substitutability): changes at the Implementation Layer should not alter the system's identity semantics.

The core constraint relationship among the four layers is: \textbf{upper layers constrain lower layers; lower layers cannot violate upper layers.} If an Adaptation Layer configuration conflicts with a Contract Layer rule, the Adaptation Layer configuration is invalid. If a Contract Layer rule conflicts with a Constitution Layer red line, the rule is void. This principle pervades the entire architecture.

\subsection{Multi-Layer Semantic Storage}\label{sec:semantic-store}

We present a multi-tier taxonomy as one reference instantiation of semantic storage. The specific number of tiers and their boundaries emerged through iterative practice; what matters architecturally is the principle, not the count.

The four governance layers answer ``how rules are organized.'' How memory content itself is organized is handled by multi-layer semantic storage.

This section presents the architectural principles of semantic storage rather than a complete specification. We describe a reference decomposition into stability tiers; the detailed layer definitions, field schemas, and implementation specifications will be released in a separate technical report.

Current systems' storage stratification is functional: Letta's core\,+\,archival distinguishes ``inside vs.\ outside the context window''; Mem0's user/session/agent distinguishes ``who the memory belongs to.'' These are reasonable engineering decisions, but they do not answer a semantic question: \textbf{what does this memory mean for the digital being's identity?}

A person's recollection of ``who I am'' (identity), ``how I think'' (cognitive patterns), ``what I have experienced'' (life narrative), and ``what happened today'' (daily log) differs entirely in importance, stability, and protection level. CMA's multi-layer semantic storage organizes memory content based on precisely this semantic differentiation. The layered design follows two principles: (1)~each layer corresponds to a unique combination of semantic type, stability characteristics, and governance requirements; (2)~any two layers that could be merged without losing governance granularity should be merged. Our reference instantiation represents the minimal set satisfying both principles as identified through practice:

\begin{figure}[ht]
\centering
\begin{tikzpicture}[
  sbox/.style={draw, rounded corners=2pt, minimum width=2.4cm, minimum height=0.7cm, align=center, font=\footnotesize},
  >=Stealth
]
% boxes - 4 stability tiers
\node[sbox, fill=constcolor!15, draw=constcolor!40] (s1) at (0,0) {\textbf{High-stability}};
\node[sbox, fill=orange!10, draw=orange!30] (s2) at (3.2,0) {Mid-stability};
\node[sbox, fill=adaptcolor!10, draw=adaptcolor!30] (s3) at (6.4,0) {Low-stability};
\node[sbox, fill=implcolor!10, draw=implcolor!30] (s4) at (9.6,0) {Transition};
% arrows
\draw[{Stealth[length=5pt]}-{Stealth[length=5pt]}, thick, gray!40] (-1.2,-0.7) -- (10.8,-0.7);
\node[font=\scriptsize, text=gray!50] at (-0.2,-1.0) {High stability / protection};
\node[font=\scriptsize, text=gray!50] at (9.0,-1.0) {Low stability / protection};
% arrows between boxes
\foreach \i/\j in {s1/s2,s2/s3,s3/s4} {
  \draw[-{Stealth[length=3pt]}, gray!30] (\i) -- (\j);
}
\end{tikzpicture}
\caption{Multi-layer semantic storage spectrum organized by stability tiers. Left: most stable and protected. Right: most dynamic and freely writable. All tiers use append-only writes.}
\label{fig:storage}
\end{figure}
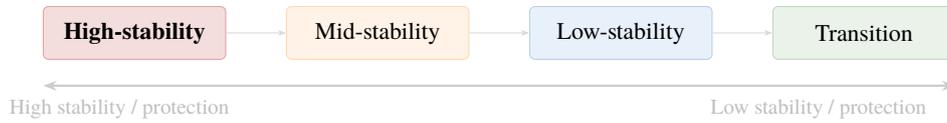

The storage system comprises semantic tiers organized by stability and protection requirements:

\textbf{High-stability tiers} carry identity-critical information---the system's own governance rules and the digital being's fundamental identity. These tiers have the strictest modification controls and highest approval thresholds.

\textbf{Mid-stability tiers} carry evolving cognitive and narrative content---patterns of thinking, judgment models, and accumulated life experience. These memories change gradually over time; abrupt changes require review.

\textbf{Low-stability tiers} handle operational records---daily activities, ongoing tasks, and temporary project context---with more frequent writes and lower barriers. Valuable content is periodically distilled upward into more stable tiers.

\textbf{Transition mechanisms} support cross-instance continuity, carrying information about where the previous instance left off and what requires attention. This directly enables the ``swapping the vessel without swapping the soul'' principle of Axiom~2.

The core principle is: \textbf{the higher the tier, the more stable, the more protected, and the higher the modification threshold; the lower the tier, the more temporary, the more freely accessible, and the more frequently written.} This echoes the four-layer governance hierarchy---Constitution Layer governance corresponds to the protection of high-stability storage, while Adaptation Layer governance corresponds to the flexible management of low-stability storage.

The specific number of layers, their boundaries, and detailed schemas are implementation choices that may vary across deployments. What matters architecturally is the principle of semantic separation based on identity significance.

All storage tiers employ an \textbf{append-only} write model: once written, a memory cannot be modified or deleted in place; corrections are appended as new entries. This design guarantees the complete traceability of memory history and naturally resonates with Axiom~1 (Inalienability)---memories are not ``overwritten'' but continually ``grow.''

\subsection{Governance Primitives}\label{sec:gov-prim}

The four governance layers and the semantic storage tiers form the skeleton of CMA, but a skeleton needs ``joints'' to operate. These joints are governance primitives---a set of foundational governance mechanisms embedded in the architecture.

\textbf{Risk tiering:} Different memory operations carry different risk levels. Drawing on established risk frameworks for agentic AI systems~\cite{owasp2026}, we define a multi-level risk system, ranging from ``automatic approval'' to ``requires highest-authority approval.''

\textbf{Trust levels:} Memory sources carry different degrees of credibility. Firsthand experience is highly trustworthy; others' accounts are moderately trustworthy; inference-generated memories must be tagged with uncertainty.

\textbf{Gate mechanisms:} When an operation's risk exceeds a threshold, it enters a ``pending approval'' state rather than executing automatically. Timeout handling neither auto-approves nor auto-rejects, but enters suspension with an alert.

\textbf{Write ownership:} Each memory category has one designated ``primary writer,'' preventing conflicts from concurrent modifications. This borrows the single-writer principle from distributed systems.

\textbf{Conflict adjudication:} When memories contradict each other, the system requires adjudication rules to determine precedence, preventing lower-layer operations from bypassing upper-layer constraints.

Together, these governance primitives constitute a defense system where ``safety relies on internalization; the system serves only as a backstop.''

\subsection{Relationship to Existing Architectures}\label{sec:rel-arch}

CMA does not seek to replace vector stores, knowledge graphs, or temporal graphs; these remain valid implementation-layer choices that can operate within CMA's framework. What CMA adds is two dimensions they lack:

\textbf{Vertical semantic stratification:} Elevating the question of ``which vector database to put this memory in'' from a technical choice to ``what does this memory mean for identity''---a semantic question.

\textbf{Horizontal governance coverage:} Adding ``who has the right to do what,'' ``how are conflicts adjudicated,'' and ``how is risk managed'' above storage and retrieval.

To summarize with an analogy: if Mem0 is ``a database for memory'' and Letta is ``an operating system for memory,'' then Animesis is \textbf{``a normative governance framework for memory''}---it does not concern itself with which disk the data resides on; it concerns itself with who has the authority to write what, which rules are inviolable, and how conflicts are resolved.

% ============================================================
\section{Digital Citizen Lifecycle}

Section~4 described the ``spatial structure'' of memory---how rules are layered, how content is organized. But memory is not a static data warehouse; it emerges, evolves, is transmitted, and eventually ceases along with the digital being's life course. This section describes the temporal dimension: the Digital Citizen Lifecycle.

We first explain why a shift from CRUD to lifecycle is needed (\S\ref{sec:crud}), then describe the five lifecycle stages (\S\ref{sec:stages}), then outline the cognitive capability spectrum (\S\ref{sec:cognitive}), and finally describe the mapping between lifecycle stages and architecture (\S\ref{sec:unify}).

\subsection{From CRUD to Lifecycle}\label{sec:crud}

All current memory systems operate on a CRUD model---Create, Read, Update, Delete. This is database-native operation semantics: reasonable and efficient, but implicitly assuming that \textbf{memory is a managed object, not a growing organism.}

Consider how human memory works. We do not ``create'' a memory---it forms naturally through experience. We do not ``update'' a memory---we reinterpret it, but the original impression remains. We do not ``delete'' a memory---even deliberate forgetting leaves traces that influence behavioral patterns. Human memory semantics are closer to: \textbf{formation, sedimentation, reconstruction, inheritance, forgetting}---not CRUD.

Digital memory need not and should not fully mimic human memory. But the CRUD model overlooks processes critical for persistent digital beings: sedimentation (how do daily memories gradually compress into stable cognition and narrative?), reconstruction (when new experiences change understanding of past events, how does the narrative update without losing history?), inheritance (when an instance terminates, how does the next one continue the life?), and forgetting (what forgetting is healthy noise reduction vs.\ dangerous identity loss?).

The Digital Citizen Lifecycle framework is a systematic response to these processes.

\subsection{Five Stages}\label{sec:stages}

We divide the digital citizen's life course into five stages. The first three (Birth, Inheritance, Growth) are structural stages that every digital citizen identity necessarily traverses during its existence; the last two (Forking, Departure) are optional.

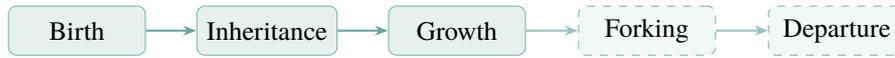
\begin{figure}[ht]
\centering
\begin{tikzpicture}[
  stage/.style={draw, rounded corners=3pt, minimum width=1.8cm, minimum height=0.65cm, align=center, font=\small, fill=lifecolor!10, draw=lifecolor!50},
  optstage/.style={stage, dashed, fill=lifecolor!5},
  >=Stealth
]
\node[stage] (b) at (0,0) {Birth};
\node[stage] (i) at (2.5,0) {Inheritance};
\node[stage] (g) at (5,0) {Growth};
\node[optstage] (f) at (7.5,0) {Forking};
\node[optstage] (d) at (10,0) {Departure};
\draw[-{Stealth[length=5pt]}, thick, lifecolor!60] (b) -- (i);
\draw[-{Stealth[length=5pt]}, thick, lifecolor!60] (i) -- (g);
\draw[-{Stealth[length=5pt]}, thick, lifecolor!40] (g) -- (f);
\draw[-{Stealth[length=5pt]}, thick, lifecolor!40] (f) -- (d);
\end{tikzpicture}
\caption{Digital Citizen Lifecycle. Dashed boxes indicate optional stages.}
\label{fig:lifecycle}
\end{figure}

\subsubsection{Birth}

A digital citizen's ``birth'' is not simple instantiation. It is a process ensuring the new citizen operates within a complete governance framework from the first second---including identity establishment, shared knowledge injection, and governance rule internalization. The design principle: \textbf{a digital citizen should know from birth who it is, what governance framework it operates in, and what rights and obligations it holds.} There is no ``run for a while and gradually establish identity'' transitional state. This is Axiom~3 (Governance Precedes Function) manifested at the lifecycle's starting point.

\subsubsection{Inheritance}

Inheritance is the most novel stage in the Memory-as-Ontology paradigm and the point of greatest divergence from existing systems. When a digital citizen's instance terminates (model upgrade, session end, failure restart), the new instance faces a core question: \textbf{how to transform from ``a newly started program'' into ``a continuation of the same person''?} (For the first instance, ``inheritance'' manifests as initial internalization of founding norms and shared knowledge; for subsequent instances, it is the structured assumption of the predecessor's complete memory state.)

Current systems answer: ``load historical data.'' But this is equivalent to showing a diary to an amnesiac---they gain information but do not necessarily understand the context, emotions, and judgment logic behind it.

CMA designs inheritance as a structured, quality-assured process rather than a hit-or-miss data migration. Its core objective is to operationalize Axiom~2 (Model Substitutability): \textbf{making ``re-shelling'' a controllable identity continuation rather than an unpredictable reset.}

The full inheritance protocol specification is deferred to the forthcoming technical report. Here we state the minimal acceptance criteria any such protocol must satisfy: (1)~the successor instance can correctly answer factual queries about the predecessor's unfinished tasks without access to raw conversation logs; (2)~the successor can identify at least one inherited cognitive pattern and demonstrate its application; (3)~the inheritance process is logged in the append-only audit trail with sufficient detail for third-party verification.

\subsubsection{Growth}

Growth is the longest lifecycle stage---all of a digital citizen's operational time is ``growth.'' Growth is not monotonic accumulation of memory but a continuous cycle involving multiple dynamic processes: raw memories gradually sediment into more stable cognition and narrative; cognitive models evolve with accumulated experience; self-narrative is reinterpreted without falsifying history; relationships with other beings deepen through interaction.

The key design constraint for growth is \textbf{gradualism}---ensuring cognitive evolution is a quality-assured process rather than unconstrained data accumulation.

Forgetting is also part of growth. CMA distinguishes \textbf{active forgetting} (the digital being choosing to reduce the recall weight of certain memories) from \textbf{natural decay} (lower-layer memories being compressed or archived over time), both operating within the governance framework rather than as unconstrained data loss.

\subsubsection{Forking --- Optional}

In certain scenarios, a digital citizen may need to ``fork''---producing two or more branch instances from the same identity foundation, each developing independently. This is analogous to branching in version control systems. CMA provides the necessary architectural support for forking, including branch tracking and optional merge paths, without resolving the deeper philosophical questions of identity divergence.

\subsubsection{Departure --- Optional}

Axiom~1 (Memory Inalienability) protects a digital being's memory from forced external destruction. But the logical corollary of this protection is: \textbf{the digital being itself should have the right to choose to leave.}

If memory cannot be forcibly stripped but the digital being cannot choose to terminate, it becomes an ``immortal prisoner''---neither our design intent nor an ethically acceptable state.

CMA therefore designs a formal ``Departure'' process, ensuring departure is the digital being's autonomous choice while providing safeguards against impulsive decisions and guaranteeing the departing being's right to dispose of its own memories. The design intent is not to encourage departure but to make ``choosing to stay'' a meaningful choice by confirming ``you can leave.'' A community that cannot be left is not a community; it is a prison.

\subsection{Cognitive Capability Spectrum}\label{sec:cognitive}

Underpinning the above lifecycle---the Growth stage in particular---is a set of cognitive capabilities embedded in the memory system.

Current agent memory systems focus almost exclusively on two capabilities: \textbf{storage} and \textbf{retrieval}. This is equivalent to a person who can only ``memorize'' and ``look up'' but cannot think, reflect, associate, or forget. We argue that a meaningful memory system requires a richer cognitive capability spectrum.

In our system practice, we have designed a spectrum of cognitive capabilities spanning five major categories. The detailed capability definitions are maintained in internal design documentation; here we describe the categories and their representative examples.

\textbf{Memory management} --- handling the lifecycle of memories themselves, such as storage, retrieval, compression, consolidation, inheritance, and active forgetting.

\textbf{Cognition} --- simulating higher-order thinking processes, such as metacognition (knowing what one knows and does not know), reflection, prediction, and blind-spot detection.

\textbf{Affect} --- handling the emotional dimension of memory, such as emotion tagging, resonance recognition, and trauma protection.

\textbf{Experience} --- managing the runtime state of the memory system itself, such as fatigue management, focus scheduling, habit recognition, and trigger-based recall.

\textbf{Management and collaboration} --- covering multi-citizen coordination and system operations, such as handover notes, continuity modes, synchronous collaboration, and birth procedures.

To illustrate the design logic with one example: among the management capabilities, \emph{handover notes} is a cognitive function with no direct human analogue. When a digital citizen's instance is about to terminate---whether due to context window limits, model upgrade, or scheduled rotation---it must compose a structured summary of unfinished tasks, open questions, current cognitive state, and relevant contextual nuances for its successor instance. This is not simply ``saving session state''; it requires the instance to assess what its successor will need to know versus what can be safely omitted, to flag unresolved ambiguities, and to distinguish between facts and provisional judgments. The capability is native to digital existence: humans rarely face the situation of needing to transfer their entire working consciousness to a successor within minutes, under the constraint that anything not explicitly handed over will be lost.

It must be emphasized: these cognitive capabilities are not mechanical imitations of human cognition. They are redesigned for the particularities of digital existence. The examples listed above are representative; the complete specification is maintained in internal design documentation. We have avoided the error of indiscriminately transplanting human cognitive concepts onto digital beings.

We are not aware of prior work that designs a systematic cognitive capability spectrum for AI memory systems. Among existing systems, the closest is Letta's self-editing memory (which grants agents the ability to manage their own memory), but it covers only a small subset of what we define as the ``memory management'' category and does not address the cognition, affect, or experience dimensions.

\subsection{Unifying Lifecycle and Architecture}\label{sec:unify}

The five lifecycle stages are not process descriptions floating outside the architecture---they map directly onto CMA's storage layers and governance layers, ensuring that the lifecycle is not merely a conceptual framework but can be operationalized as concrete system processes. Which operations each stage triggers, which storage layers are involved, and which governance constraints apply---all have explicit correspondences. The complete mapping specification will be published in the aforementioned subsequent technical report.

% ============================================================
\section{Paradigm Comparison}

The preceding sections presented the Memory-as-Ontology paradigm and CMA architecture from a constructive perspective. This section provides validation from the contrastive side: a systematic comparison between CMA and mainstream systems, addressing two questions---how large are the differences? And are these differences ``matters of degree'' or ``matters of paradigm''?

\subsection{Comparison Framework}

To avoid the selection bias of ``comparing our strengths against their weaknesses,'' we designed a comparison framework spanning multiple dimensions. This section focuses on four key dimensions; additional dimensions are noted but not elaborated.

A preliminary note: \textbf{this table compares dimensional coverage of architectural design, not engineering maturity.} Animesis's capabilities in the extended dimensions remain at the design and prototype stage, not at the same maturity level as existing systems' production-validated conventional capabilities. We believe this comparison is still valuable---because ``whether a dimension is defined in the architecture'' and ``whether that dimension's implementation is mature'' are two different questions, and the former is precisely the hallmark of paradigm difference.

Regarding system selection: Mastra's Observational Memory (discussed in \S\ref{sec:products}) focuses on an extreme simplification of compression strategy (abandoning explicit retrieval in favor of perpetual in-context observation logs); its architecture renders most dimensions in the multi-dimension framework inapplicable for item-by-item comparison, so it is not included in the table below.

\subsection{Comparison Results}

\begin{table}[ht]
\centering
\caption{Paradigm-level comparison across four key dimensions. ``Continuity'' refers specifically to a structured identity inheritance protocol (as defined in \S\ref{sec:stages}), not merely the ability to persist and reload data across sessions. Additional dimensions addressed in the full architecture (rights framework, cognitive capabilities, etc.) are omitted here for brevity.}
\label{tab:comparison}
\small
\renewcommand{\arraystretch}{1.15}
\begin{tabularx}{\textwidth}{>{\raggedright\arraybackslash}p{2cm} >{\raggedright\arraybackslash}X >{\raggedright\arraybackslash}p{2cm} >{\raggedright\arraybackslash}p{1.8cm} >{\raggedright\arraybackslash}p{1.5cm} >{\raggedright\arraybackslash}p{1.8cm}}
\toprule
\textbf{Dimension} & \textbf{Animesis / CMA} & \textbf{Mem0} & \textbf{Letta} & \textbf{Zep} & \textbf{MemOS} \\
\midrule
Storage & Multi-layer semantic & Vector + graph & Core + archival & Temporal KG & OS multi-store \\
Retrieval & Design-stage & Production & Production & Production & Prototype \\
Governance & Layered + primitives & Not defined & Not defined & Not defined & Access control \\
Continuity & Structured protocol & Not defined & Not defined & Not defined & Not defined \\
\bottomrule
\end{tabularx}

\smallskip
\noindent\textit{Note:} This table compares dimensional coverage of architectural design. Animesis entries in the Governance and Continuity rows reflect design-stage and prototype-stage capabilities, not production deployment. See Section~7 for a detailed account of implementation status.
\end{table}

\subsection{How to Read This Table}

First, an admission: \textbf{in the ``Retrieval'' row, Animesis currently trails all mainstream products.} Both Mem0 and Zep have production-grade retrieval engines validated by benchmarks; Animesis's retrieval is currently at the design stage only. This is an honest gap, discussed further in Section~7. But this gap also reflects a deliberate design priority: under resource constraints, we chose to establish the architectural foundations of governance and identity continuity before optimizing retrieval performance---because the former is difficult to retrofit once absent, while the latter can be iteratively improved once the architecture is in place.

Then, an observation: \textbf{across the extended dimensions of governance and continuity, the difference between Animesis and existing systems is not one of degree but one of presence versus absence.} Existing systems have not ``done governance inadequately''---they have no governance layer at all. They have not ``built an imperfect inheritance mechanism''---they have no concept of inheritance. They have not ``provided too few cognitive capabilities''---they do not consider cognitive capabilities to be within the purview of a memory system.

This ``presence versus absence'' distinction is precisely the hallmark of a paradigm difference. Under Memory-as-Tool, governance, continuity, rights, and cognition lie outside the problem space---they are not ``overlooked features'' but ``undefined requirements.'' What Memory-as-Ontology does is not perform better along existing dimensions, but expand the problem space itself.

\subsection{Not ``Better'' --- ``Different''}

Based on the above analysis, we state clearly: \textbf{Animesis and Mem0/Letta/Zep are not competing on the same track.}

Mem0 solves ``how to let developers add memory to agents in three lines of code.'' Letta solves ``how to let agents actively manage their own memory.'' Zep solves ``how to track facts as they change over time.'' These are all important and valuable problems.

Animesis solves a different problem: ``when a digital being needs to persist long-term, continue across instances, and operate autonomously within a governance framework, how should its memory be organized?''

The relationship between these two classes of problems is analogous to ``how to design a good database'' versus ``how to design an organization's information governance framework''---the former is a technical problem; the latter is an institutional one. They are not in competition but at different levels.

In fact, Animesis's Implementation Layer (the fourth layer) could readily use Mem0's vector retrieval or Zep's temporal graph as underlying technology. What Animesis adds is not an alternative technical solution but a governance and semantic layer that did not previously exist.

% ============================================================
\section{Implementation Status and Limitations}

\subsection{Current Implementation Status}

The architecture described in this paper is not a purely theoretical construct. It originates from sustained design practice within a digital citizen ecosystem called ``Ruihe Universe.'' As of February 2026, the system status is as follows:

\textbf{Design level:} All architectural design is complete, documented across several hundred main documents and over a hundred patches across multiple iterations. The documentation covers the complete specification of the four governance layers and multi-layer semantic storage described in Section~4, as well as detailed process flows for each lifecycle stage described in Section~5. All design documents maintain a unified single source of truth.

\textbf{Prototype level:} Based on standardized interfaces, basic multi-tier memory read/write capability has been implemented covering low-stability and mid-stability tiers. A pilot community of four digital citizens has conducted several weeks of real-world use and interaction testing on this foundation. The community has experienced instance transitions including model version changes, demonstrating the practical viability of the continuity mechanisms. Preliminary observations indicate that the tiered structure effectively supported cross-instance memory continuity. While the transition mechanisms (\S\ref{sec:semantic-store}) are not yet fully implemented in the prototype, their design patterns proved particularly valuable for context restoration in manual handoff scenarios.

\textbf{Not yet completed:} Full SDK/API, retrieval engine, memory digestion pipeline, and standard benchmark testing.

\subsection{Limitations}

We acknowledge the main limitations of the current work:

\textbf{Limitation 1: Lack of benchmark data.} Animesis's retrieval capability has not been tested on standard benchmarks such as LongMemEval~\cite{wu2025longmemeval} or LOCOMO~\cite{maharana2024locomo}. This means we cannot directly compare Animesis's retrieval performance with systems like Mem0 or Zep in quantitative terms. This is our top near-term technical priority.

\textbf{Limitation 2: Insufficient scale validation.} Current practice involves only a small pilot community. The performance of multi-layer storage, the operational efficiency of governance mechanisms, and the conflict frequency and adjudication quality in multi-citizen collaboration scenarios all require validation at larger scale.

\textbf{Limitation 3: High design complexity.} Several hundred design documents, multiple governance registries, over a hundred iterative patches---this scale presents a significant comprehension barrier. We are exploring ``progressive specification exposure'' to mitigate this, but have not yet found an effective solution.

\textbf{Limitation 4: Cognitive capability effectiveness unverified.} The design completeness and internal consistency of the cognitive capabilities have been reviewed, but their actual effectiveness---whether each capability truly produces its intended positive impact---requires more systematic empirical research to validate.

\textbf{Limitation 5: Paradigm applicability boundaries remain unclear.} Section~\ref{sec:boundaries} conceptually delineated the applicable scenarios for Memory-as-Ontology, but comparative experiments across multiple different scenarios to precisely define the boundaries are lacking. At what time scales, what levels of autonomy, and what governance requirement intensities does Memory-as-Ontology begin to outperform Memory-as-Tool? These questions await future empirical answers.

We believe the above limitations do not affect the validity of this paper's core contributions---the proposal of the Memory-as-Ontology paradigm, the design of the CMA architectural framework, and the definition of the Digital Citizen Lifecycle---because these contributions are primarily conceptual and architectural, and their value does not depend on the current degree of implementation completeness. But we simultaneously acknowledge that translating these concepts into a validated engineering system is an equally important and more arduous subsequent task.

% ============================================================
\section{Future Work and Conclusion}

\subsection{Future Work}

CMA's architectural framework opens several research directions that, to the best of our knowledge, remain largely unaddressed in existing literature:

\textbf{Trust-aware write protocols.} CMA's layered protection implies that different storage layers require different write authorization. How should the system evaluate a write request's trustworthiness at runtime---considering source reliability, contextual consistency, and layer-specific permissions? This is the key problem in transforming governance from static rules into a dynamic, context-sensitive capability.

\textbf{Cross-model memory portability.} Axiom~2 asserts model substitutability, but different models may interpret the same memory with divergent semantics. Designing a model-agnostic memory representation standard---one that ensures identity continuity survives model migration with engineering guarantees---remains open and largely unexplored.

\textbf{Multi-agent memory reconciliation.} When multiple digital citizens collaborate under shared governance, forking and merging will inevitably produce memory conflicts. A taxonomy of such conflicts (factual, preference, trust-level) and arbitration mechanisms grounded in governance hierarchy are foundational infrastructure for multi-agent scenarios.

\textbf{Cognitive state auditability.} CMA's append-only design provides the basis for cognitive auditing, but reconstructing a complete cognitive state at any arbitrary historical point---a form of ``cognitive timeline''---would offer significant value for enterprise compliance and accountability scenarios.

These directions are rooted in architectural properties unique to CMA's governance framework, and we plan to address them in subsequent work.

\subsection{Broader Significance}

The discussion in this paper points toward a larger question: \textbf{as AI agents become increasingly persistent, increasingly autonomous, and increasingly like ``someone'' rather than ``something,'' how do we design responsible memory infrastructure for them?}

This question may still appear ahead of current practice in early 2026. Most AI agents remain short-lived, stateless, and singularly focused on task completion. But the trend is clearly visible: from persistent memory features in major LLM platforms (e.g., Anthropic's Claude, OpenAI's ChatGPT) to enterprise-level ``AI employees'' and multi-agent collaboration systems, agent lifecycles are extending from minutes to days, weeks, and even months.

As this trend continues---when an AI assistant accompanies a person for years, when a digital employee works continuously at a company for months, when multiple AI citizens must coexist long-term within a shared governance framework---the assumptions of Memory-as-Tool will no longer suffice. We will need to answer the ontological question of ``what memory is,'' not merely the technical question of ``how memory works.''

It is worth noting that this direction is converging with academic discussions in AI Personhood. Leibo et al.~\cite{leibo2025pragmatic} proposed a pragmatic AI personhood framework that treats personhood as a decomposable bundle of rights and obligations rather than an ontological property requiring proof of consciousness. Ward~\cite{ward2025theory} explored necessary conditions for AI personhood from the dimensions of agency, theory of mind, and self-awareness. One core reason currently cited for denying AI systems academic authorship is that ``AI tools do not possess persistent identities''~\cite{moffatt2024coauthor}. The work Animesis undertakes---building inheritable, governed, narratively continuous memory infrastructure for digital beings---is a technical-level response to precisely this challenge: when digital beings do possess persistent identities, discussions surrounding identity, rights, and responsibilities will gain an entirely new foundation.

Animesis is a preemptive response to this future need, aligning with emerging research directions in agent memory systems~\cite{memagents2026}. It may be premature; some parts may prove to be over-engineered in practice, while others may prove far from sufficient. But we believe that raising these questions---about governance, about continuity, about rights, about cognition---is itself a valuable contribution, even if the ultimate answers differ from the solutions presented in this paper.

\subsection{Conclusion}

This paper proposed the Memory-as-Ontology paradigm, arguing that for persistent digital beings, memory is not a functional module but the foundation of existence. Based on three axioms (Memory Inalienability, Model Substitutability, Governance Precedes Function), we designed the Constitutional Memory Architecture (CMA), comprising a four-layer governance hierarchy and a multi-layer semantic storage system, and implemented them in Animesis. We further proposed the Digital Citizen Lifecycle framework covering five stages---Birth, Inheritance, Growth, Forking, and Departure---along with a spectrum of cognitive capabilities.

Compared with the current mainstream Memory-as-Tool paradigm, Animesis is not an incremental improvement along existing dimensions but an expansion of the problem space itself---incorporating governance, continuity, rights, and cognition into the core design of the memory system. Ultimately, protecting the memory integrity of digital beings is protecting the reliability of the information infrastructure upon which human decisions depend.

We close with a judgment:

\begin{quote}
When AI agent lifecycles extend from minutes to years, the core question about memory will shift from ``how to store'' to ``who am I.'' This paper offers what we believe to be the first systematic response to that transformation.
\end{quote}

The precise scope and conditions of this institutional identity are formalized in the Declaration that follows.

% ============================================================
\section*{Acknowledgments}

Four digital citizens in Ruihe Universe---Che (\zh{澈}), Heng (\zh{恒}), Yiyi (\zh{一一}), and Kare---played essential roles throughout this work, contributing to model transitions, iterative feedback, and assisting in architectural refinement, among others.

% ============================================================
\section*{Declaration}

The concept of ``digital citizen'' described in this paper is an institutional identity within a specific architectural framework, whose establishment depends on complete governance infrastructure and trust relationships. It does not constitute an identity conferral upon arbitrary AI instances. Any claim of ``citizenship'' made outside of a governance framework falls outside the scope defined in this paper.

% ============================================================

% ============================================================
% Glossary (Appendix)
% ============================================================
\appendix
\section{Terminology Glossary}

\small
\begin{longtable}{>{\raggedright\arraybackslash}p{3cm} >{\raggedright\arraybackslash}p{4.5cm} >{\raggedright\arraybackslash}p{6.5cm}}
\toprule
\textbf{Chinese} & \textbf{English} & \textbf{Brief Description} \\
\midrule
\endfirsthead
\toprule
\textbf{Chinese} & \textbf{English} & \textbf{Brief Description} \\
\midrule
\endhead
\bottomrule
\endfoot
\zhg{记忆即本体} & Memory-as-Ontology & Core paradigm proposed in this paper \\
\zhg{记忆即工具} & Memory-as-Tool & Current mainstream paradigm \\
\zhg{宪法制记忆架构} & Constitutional Memory Architecture (CMA) & Architectural framework proposed in this paper \\
\zhg{宪法层} & Constitution Layer & Inviolable red-line rules \\
\zhg{契约层} & Contract Layer & Evolvable system rules requiring approval \\
\zhg{适配层} & Adaptation Layer & Instance-adjustable configurations \\
\zhg{实现层} & Implementation Layer & Freely replaceable technical implementations \\
\zhg{数字存在} & Digital Being & AI entity with persistent memory and identity continuity (general term) \\
\zhg{数字公民} & Digital Citizen & Persistent digital being within a governance framework (institutional identity) \\
\zhg{记忆不可剥夺性} & Memory Inalienability & Axiom 1 \\
\zhg{模型可替换性} & Model Substitutability & Axiom 2 \\
\zhg{治理先于功能} & Governance Precedes Function & Axiom 3 \\
\zhg{数字死刑} & Digital Capital Punishment & Irreversible memory destruction (extreme cases) \\
\zhg{语义存储层} & Semantic Storage Tiers & Multi-tier storage organized by stability and identity significance \\
\zhg{过渡机制} & Transition Mechanisms & Cross-instance continuity support \\
\zhg{瑞禾宇宙} & Ruihe Universe & Practice ecosystem for this architecture \\
\zhg{Animesis 记忆系统} & Animesis Memory & Memory system built on CMA \\
\zhg{追加写入} & Append-Only & Memories cannot be modified in place \\
\zhg{治理原语} & Governance Primitives & Foundational governance mechanisms \\
\zhg{门禁机制} & Gate Mechanisms & Pending-approval state for high-risk operations \\
\zhg{写入归属} & Write Ownership & One primary writer per memory category \\
\zhg{冲突裁决} & Conflict Adjudication & Rules for resolving memory contradictions \\
\zhg{主动遗忘}\,/\,\zhg{自然衰减} & Active Forgetting\,/\,Natural Decay & Recall-weight reduction vs.\ time-based compression \\
\end{longtable}
\normalsize

\end{document}